\documentclass[acmsmall, nonacm]{acmart}
\renewcommand\footnotetextcopyrightpermission[1]{} 
\AtBeginDocument{%
  \providecommand\BibTeX{{%
    \normalfont B\kern-0.5em{\scshape i\kern-0.25em b}\kern-0.8em\TeX}}}

\setcopyright{acmlicensed}
\copyrightyear{2024}
\acmYear{2024}
\acmDOI{XXXXXXX.XXXXXXX}

\usepackage[utf8]{inputenc}

\usepackage[T1]{fontenc}

\usepackage{microtype}

\usepackage{bm,mathtools,color}

\usepackage[normalem]{ulem}
\usepackage{enumitem}

\usepackage{booktabs}
\usepackage{multirow}
\usepackage{subcaption}
\usepackage{tabularx}
\usepackage{multicol}
\usepackage{multirow}
\usepackage{colortbl}
\usepackage{hhline}
\usepackage{stfloats}
\usepackage{float}

\newcolumntype{Y}{>{\centering\arraybackslash}X}

\usepackage{booktabs}
\usepackage{multirow}
\usepackage{graphicx}

\usepackage{pifont}
\usepackage{amsmath}
\usepackage{ifthen}
\usepackage{dsfont}

\usepackage{listings}
\usepackage{xcolor}
\usepackage{ltablex}
\usepackage{longtable}
\usepackage{multicol}

\usepackage{tikz}
\usetikzlibrary{bayesnet}
\usepackage{subfloat}
\usepackage[linguistics]{forest}

\usepackage{algorithm}
\usepackage{algpseudocode}

\algrenewcommand\algorithmicrequire{\textbf{Input:}}
\algrenewcommand\algorithmicensure{\textbf{Output:}}
\algnewcommand{\LeftComment}[1]{\Statex \(\triangleright\) #1}

\usepackage{caption}

\usepackage{cancel}

\usepackage{xspace}

\newcommand{\prob}[2][]{p\ifthenelse{\not\equal{}{#1}}{_{#1}}{}(#2)} 
\newcommand{\expect}[2][]{\text{\bf E}\ifthenelse{\not\equal{}{#1}}{_{#1}}{}\!\left[#2\right]}
\newcommand{\var}[2][]{\text{\bf Var}\ifthenelse{\not\equal{}{#1}}{_{#1}}{}\!\left[#2\right]}

\newcommand{\bx}{\mathbf{x}}

\newcommand{\by}{\mathbf{y}}

\newcommand{\name}{NetFlowGen}

\newcommand{\xatu}{EarlyDetect}

\newcommand{\bfsection}[1]{\vspace{0.3cm}\noindent\textbf{#1.}}
\newcommand{\bfsectionn}[1]{\vspace{0.3cm}\noindent\textbf{#1}}

\newcommand{\presec}{\vspace{-0.00in}}
\newcommand{\postsec}{\vspace{-0.00in}}
\newcommand{\presub}{\vspace{-0.00in}}
\newcommand{\postsub}{\vspace{-0.00in}}

\newcommand{\prefig}{\vspace{-0.00in}}
\newcommand{\postfig}{\vspace{-0.00in}}
\newcommand{\prefigcaption}{\vspace{-0.00in}}
\newcommand{\postfigcaption}{\vspace{-0.00in}}

\newcommand{\pretable}{\vspace{-0.00in}}
\newcommand{\posttable}{\vspace{-0.00in}}
\newcommand{\pretablecaption}{\vspace{-0.00in}}
\newcommand{\posttablecaption}{\vspace{-0.00in}}

\newcommand{\draftonly}[1]{#1}
\renewcommand{\draftonly}[1]{}

\usepackage{xcolor}
\definecolor{red}{rgb}{0.74,0.08,0.10}
\definecolor{green}{rgb}{0.26,0.49,0.18}
\definecolor{blue}{rgb}{0.22,0.53,0.75}

\newcommand{\ensuretext}[1]{#1}

\newcommand{\marker}[2]{\ensuremath{^{\textsc{#1}}_{\textsc{#2}}}}
\newcommand{\authorcomment}[4]{\draftonly{\ensuretext{\textcolor{#3}{[\marker{#1}{#2} #4]}}}}

\newcommand{\joe}[1]{\authorcomment{J}{Z}{orange!80}{#1}}
\newcommand{\woojeong}[1]{\authorcomment{W}{K}{blue!80}{#1}}

\newcommand{\minlan}[1]{\authorcomment{M}{Y}{blue}{#1}}

\newcommand{\notes}[1]{\draftonly{{\color{red}#1}}}
\newcommand{\outline}[1]{\draftonly{{\color{violet}\noindent [Outline] #1}}}

\begin{document}


\title{NetFlowGen: Leveraging Generative Pre-training for Network Traffic Dynamics}

\author{Jiawei Zhou}
\authornote{First three authors contributed equally to this research.
Work done at Harvard University.
}
\email{jiawei.zhou.1@stonybrook.edu}
\orcid{https://joezhouai.com}
\affiliation{%
  \institution{Stony Brook University}
  \streetaddress{}
  \city{}
  \state{}
  \country{USA}
  \postcode{}
}

\author{Woojeong Kim}
\authornotemark[1]
\email{wk247@cornell.edu}
\orcid{https://www.wkim.info/}
\affiliation{%
  \institution{Cornell University}
  \streetaddress{}
  \city{}
  \country{USA}}

\author{Zhiying Xu}
\authornotemark[1]
\email{zhiyingxu@g.harvard.edu}
\orcid{https://xuzhiying9510.github.io/}
\affiliation{%
  \institution{Harvard University}
  \city{}
  \country{USA}
}

\author{Alexander M. Rush}
\email{arush@cornell.edu}
\orcid{https://rush-nlp.com/}
\affiliation{%
 \institution{Cornell University}
 \streetaddress{}
 \city{}
 \state{}
 \country{USA}}

\author{Minlan Yu}
\email{minlanyu@g.harvard.edu}
\orcid{https://minlanyu.seas.harvard.edu/}
\affiliation{%
  \institution{Harvard University}
  \streetaddress{}
  \city{}
  \state{}
  \country{USA}}

\renewcommand{\shortauthors}{Zhou, et al.}


\begin{abstract}

Understanding the traffic dynamics in networks is a core capability for automated systems to monitor and analyze networking behaviors, reducing expensive human efforts and economic risks through tasks such as traffic classification, congestion prediction, and attack detection.
However, it is still challenging to accurately model network traffic with machine learning approaches in an efficient and broadly applicable manner.
Task-specific models trained from scratch are used for different networking applications, which limits the efficiency of model development and generalization of model deployment.
Furthermore, while networking data is abundant, high-quality task-specific labels are often insufficient for training individual models.
Large-scale self-supervised learning on unlabeled data provides a natural pathway for tackling these challenges.
We propose to \textit{pre-train} a general-purpose machine learning model to capture traffic dynamics with only traffic data from NetFlow records, with the goal of \textit{fine-tuning} for different downstream tasks with small amount of labels.
Our presented {\name} framework goes beyond a proof-of-concept for network traffic pre-training and addresses specific challenges such as unifying network feature representations, learning from large unlabeled traffic data volume, and testing on real downstream tasks in DDoS attack detection.
Experiments demonstrate promising results of our pre-training framework on capturing traffic dynamics and adapting to different networking tasks.

\end{abstract}

\maketitle

\presec\section{Introduction}
\label{sec:introduction}\postsec

\notes{
Pretraining is useful

Networking is challenging for pretraining (diff between networking and NLP)
- raw feature representation for ML/pre-training, 
- heterogeneous features: discrete vs continuous; their distributions, ----> need good representation
- much scale variance: range of feature values is very large, IP: $2^32$. #packets: from 0 pkt to billions of pkts
- networks have continuous features that 

New opportunities of pretraining:
- generalize to different tasks (in hotnets)
- Requires fewer clean labels

Design decision: 
-General model: do not customize models for networking
1. - Decoder is the trend, better than encoding schemes
2. easily changable to future new bettxrer models
3. good at any types (heterageneity) of traffic features
4. Expandable to other types of data .. (like text input)
- Do not customize data for scale reasons 
1. Do not customize feature selection
2. Many techniques on scaling memory, features. 

- Pretraining task: predict next traffic features; general basis for many advanced tasks
1. multitasking: predicting different trime step  features...different with NLP

Networking brings opportunities

Attacks

Labeling issues in networks:
\begin{itemize}
    \item Labels expensive to acquire
    \item Labels are not reliable (or very noisy)
\end{itemize}

Issues with task-specific fine-tuning:
\begin{itemize}
    \item Expensive in data collection, algorithm design, etc.
    \item Models do not generalize. Features are limited in usage for broader domains.
\end{itemize}

Issues with pre-training in networks:
\begin{itemize}
    \item Only preliminary studies, in limited setup.
    \item Does not include complex features, and their representation for machine learning models.
    \item Data scale is limited, and only on simulated data.
    \item Tasks considered are very limited in both the importance and application in the wild. More of a proof of concept, instead of real applications
\end{itemize}

In this work, we address the previous issues in networks, and particularly apply pre-training techniques for network data.
We ... large scale and real dataset for the first time, experimented on a real task of DDoS attack detection.
We also detail how to effectively represent network features in the machine learning model.
}

\outline{

importance and issues in networking tasks

why use pre-training? (labeling issues, massive data)

what challenges are there in network pre-training?

what do we do, and out contributions

}


Comprehensive understanding and rigorous modeling of traffic dynamics are important for achieving accurate and reliable network service and management. With the capture of rich network data, 
recent years have seen a surge in applying machine learning (ML) techniques in a wide range of networking tasks, including congestion prediction \cite{poularakis2021generalizable} and control \cite{abbasloo2020classic, jay2019deep, nie2019dynamic, winstein2013tcp}, packet \cite{liang2019neural, zhang2021fast} and traffic classification \cite{casino2019hedge, meng2022packet, shen2019encrypted, zhang2020autonomous, zheng2020learning}, traffic optimization \cite{chen2018auto}, performance prediction and estimation \cite{manousis2020contention, zhang2021mimicnet}, network traffic generation \cite{ring2019flow, yin2022practical}, and attack detection \cite{fu2018intelligent, kharraz2018surveylance, wu2020network, zhou2020automating}.
As a paradigm, ML models achieve strong performance for network applications \cite{boutaba2018comprehensive}, usually through supervised learning from in-task labeled datasets.

Nevertheless, while individual ML models are effective for specific applications, it is challenging to transfer learned knowledge and network representations from task to task to achieve better efficiency and generalization.
Apart from the repetitive cost of data collection and model development customized for different tasks, the inadequacy of high-quality ground truth labels poses a unique challenge in obtaining reliable models \cite{rathore2018semi}.
Although there naturally exists an ample amount of networking data \cite{hong2009traffic, kornexl2005building, wang2021examination}, most raw data are unlabeled. Acquiring task-specific labels with different learning purposes is a very expensive process. For example, flow-based traffic data might contain billions of flows and packets, for training a reliable attack detection ML model, manually labeling all the data with high quality is extremely time-consuming even for experts in security \cite{ring2019flow}.

The recent success of \textit{self-supervised learning} from large-scale unlabeled data in natural language processing and related fields motivates us to explore its applications in network traffic dynamics modeling.
The self-supervised learning framework contains two phases: a \textit{pre-training} phase that learns general feature representations from a large amount of unannotated data, and a \textit{fine-tuning} phase that adapts the general model for particular purposes with a much smaller set of task-specific labels. This \textit{transfer learning} process uses knowledge encapsulated in the general pre-trained model to transfer to a variety of downstream tasks without redesigning the backbone model architectures and training model parameters from scratch.
For network applications, the expensive data labeling issues along with the existence of massive unlabeled networking data make the pre-training paradigm a natural fit as a potential solution for efficient and generalizable ML systems \cite{le2022rethinking}.

\minlan{Propose your idea here: we propose NetFlowGPT, a framework for
generative pre-training on NetFlow [10] traffic data....}

\minlan{Then talk about each challenge of networking and how you address each challenge}


In this work, we propose {\name}, a framework for \textit{generative} pre-training on NetFlow \cite{claise2004cisco} traffic data.
Although pre-training frameworks offer potential opportunities for networks, only a handful of works have explored them, primarily discussing benefits \cite{le2022rethinking} or providing proof of concepts under small-scale simulation setups \cite{dietmuller2022new}. Challenges persist in developing operational systems with pre-training techniques applied to real-world network traffic data.
These include handling the complexities of diverse traffic features, designing effective training objectives with proper model architectures, assembling real-world traffic pre-training datasets at scale, and establishing benchmarks for testing on realistic downstream tasks \cite{le2022rethinking}.

In response to these challenges, we focus on pre-training on flow-based traffic from ISPs (Internet Service Providers), which are in a unique position with access to all or majority of incoming traffic from customers served, employing opportunities for larger data collection and broader usability.
While traffic dynamics superficially resemble sequence modeling problems in other fields, there are substantial differences in the raw elements of traffic streams from other data such as language. We design a general feature embedding pipeline that incorporates rich and heterogeneous traffic features from NetFlow data, including IPs, timestamps, ports, transport protocols, packets, bytes, etc., and can be extended for new features. All features are cast into a common space within the pre-trained model, encompassing both incoming and outgoing traffic of an IP node.


For model and learning, we use a pre-training task of predicting the next time-step based on the history of traffic information. This generative process relies solely on raw traffic data extracted from NetFlow records \cite{claise2004cisco, claise2008specification}, without any labels. We adopt the Transformer architecture \cite{vaswani2017attention}, commonly used in NLP applications, specifically a decoder-only architecture like OpenAI GPT \cite{radford2018improving}. The resulting model learns to simultaneously generate various traffic features, leveraging diverse signals and feature interactions to acquire comprehensive knowledge representations for downstream tasks.

\minlan{This is already about evaluation. We tried our idea on NetFlow data from a large ISP...Zhiying may rewrite the whole paragraph as the evaluation paragraph in intro}
\joe{set up the tone: as the first step to build a realistic pre-training model for network traffic applications, etc.}

 For dataset construction, we, for the first time, utilize a large-scale network traffic corpus collected from a large ISP over three consecutive months for pre-training. We then fine-tune the model on a real task of DDoS early detection to showcase the advantages of large-scale pre-training.
Our experiments consider 86 traffic features during the pre-training stage. The resulting models exhibit improved performance in various attack detection tasks for network security, even with limited labeled data.
Our preliminary results with {\name} demonstrate the promising benefits of incorporating self-supervised pre-training frameworks into networking research. The general design of our approach can be further extended and tested in diverse environments and applications.

\presec\section{Challenges and opportunities}
\label{sec:challenges}\postsec

\notes{
NLP is successful with pre-training. can we do similar in networks. how it is different.

- feature: networking
- model: general model - we do not want to change model; not customized models, data centeric; there are not so much need to change the model, the model 
- tasks: what are the tasks are, labeling issues, etc.

major section is this section 2: meant to educate ppl what are the advancement are there in NLP, what are the challenges, what should be the potentially right way to move forward

==============

next section:

feature handling

model choice

pre-training task

Title: Net-GPT etc.

Some background here.

Pre-training and transfer learning

network attack detection

noisy and inadequate label for attacks

pre-training on network data

}

\outline{

A bit of background in pre-training, transfer learning, popular models

Pre-training helps NLP. How is networks different from NLP?
\begin{itemize}
    \item[-] Can we achieve similar success?
    \item[-] What are the challenges arising from these differences?
\end{itemize}

What could be the "right" or better way of doing pre-training in networks?
\begin{itemize}
    \item[-] feature: networking heterogenous features, we want to include them with uniformed way, and also generalizability
    \item[-] model: general model that can be similar to other fields such as NLP; decoder only
    \item[-] data and task: more data in networks, tasks need to carry good dynamics, less benchmarks in standard tasks, 
\end{itemize}

}


\begin{figure}
    \centering
    \prefig\includegraphics[width=0.9\linewidth]{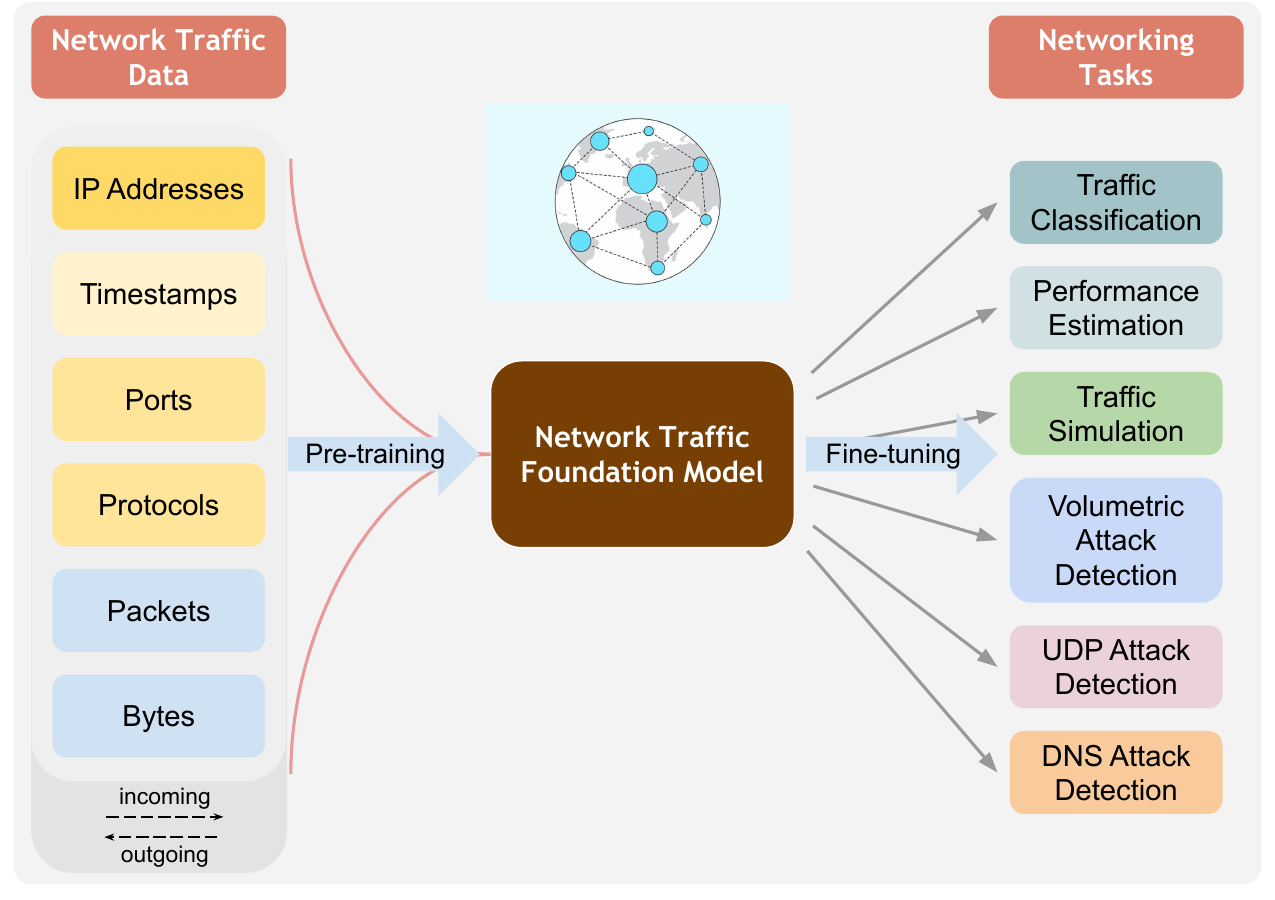}\postfig
    \prefigcaption\caption{A desirable vision of a network foundation model 
    that captures comprehensive traffic dynamics and the same model can be adapted to various downstream networking tasks.
    }
    \label{fig:vision-network-foundation}\postfigcaption
\end{figure}

The pre-training paradigm has greatly reshaped the landscape of natural language processing and related fields. The core idea is to first pre-train a self-supervised model, with large amount of unlabeled data and then transfer the model to new tasks with minimal supervision. Systems start with a common Transformer \cite{vaswani2017attention} neural network architecture.
Since the framework was introduced, pre-trained models have improved continuously: from BERT \cite{devlin2018bert} and RoBERTa \cite{liu2019roberta}, to T5 \cite{raffel2020exploring} and BART \cite{lewis2019bart}, and to GPT \cite{radford2018improving} and PaLM \cite{chowdhery2022palm}. 
Language processing systems have advanced on these models, also referred to as a foundation model \cite{bommasani2021opportunities}, along with the massive scaling of data and model size.
Beyond NLP, foundation models have found applications in different data domains such as graphs \cite{liu2023towards, zhou2021structure, zhou2023generating}, computer vision \cite{dosovitskiy2020image, gupta2023cliptrans, lisocialgpt}, health \cite{luo2022biogpt}, and finance \cite{wu2023bloomberggpt}.



However, in the field of networks, these approaches have not been well explored. Can we achieve similar success in networking research with the vision depicted in Figure~\ref{fig:vision-network-foundation}, and how can we design and train such models for networks?

\minlan{Add a bolded sentence at the beginning of each paragraph to highlight the key challenges}
\bfsection{Feature Diversity}
Networking data is a challenge for pre-training. Current state-of-the-art Transformer models by default operate on a sequence of univariate tokens as input,\footnote{These tokens can be text words, image patches, etc.} which take discrete values from a finite vocabulary. Networking data is both multivariate and heterogeneous. 
There are different fields composing the traffic information, including IP addresses, ports, transport protocol, timestamps, packets, bytes, etc.\joe{check to use number of packets, in a flow or? we focus on traffic volume, more macro scope}
Some fields are continuous such as packets and bytes, whereas others are categorical such as ports and protocol.
Moreover, the distributions of raw feature values vary drastically. While possible values of protocols are limited, the values of the number of packets in the traffic flow range from zero to billions.
A comprehensive and effective pre-trained model needs to incorporate each of the raw traffic features to ensure broad generalization, and represent them in a unified space to learn their interactions, regardless of feature types and distributions.
\minlan{Can you give some evidence like if we naively feed in network data into GPT, the accuracy is very low. Right now, the arguments are pretty high-level, it would be good to provide a quantified view }

\bfsection{Pre-Training Objective}
We also need to consider how to design the best pre-training task for networks.
Network traffic shares similarities with other modalities due to its sequential structure. This leads us to believe that models with \textit{next-step} pre-training prediction task, also known as decoders, can be readily adapted to networks for learning traffic dynamics. We note that this choice differs from previous work \cite{dietmuller2022new} which explored preliminary network pre-training with bidirectional encoders. We believe there are several benefits to next-step prediction including the ability to handle streaming data and take advantage of developments in general pre-training, such as memory optimization and model scaling for the model architectures. However, even with a decoder there do exist difficulties in a proper formulation of the traffic sequences from raw data such as in NetFlow records.


\bfsection{Data Collection \& Curation}
The success of unlabeled data pre-training is highly dependent on large amount of clean data, primarily for the pre-training stage to obtain high-quality representations of network dynamics but also for downstream tasks to fairly benchmark and improve model performances. Due to the high variance of data properties for different network services and the heterogeneous nature of network information, it is challenging to curate well-balanced datasets for general-purpose pre-training. For benchmark tasks, ground truths are often impossible to acquire, such as true attacks in network security, expert labeling is expensive, and there also exist privacy concerns.

Nevertheless, the abundance of network traffic data provides a natural testbed for pre-trained models to capture traffic dynamics. With many challenges unresolved, it brings great opportunities for networking research. We explore some concrete resolutions to connect network problems with performant modeling techniques in NLP, with the hope to shed some light on the future development of general network traffic foundation models.

\presec\section{Method}
\label{sec:method}\postsec

\begin{figure}
    \centering
    \prefig\includegraphics[width=\linewidth]{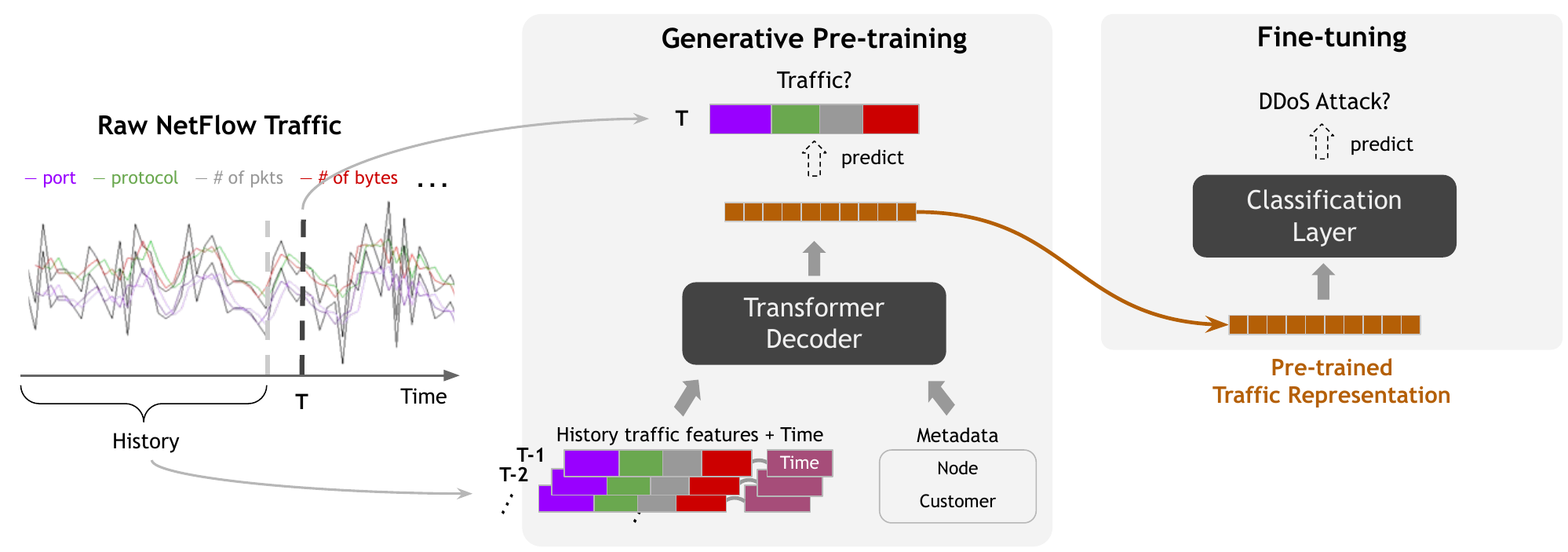}\postfig
    \prefigcaption\caption{{\name} generative pre-training framework. The framework consists of two parts: generative pre-training and fine-tuning. The objective of pre-training is to predict the traffic of time-step $T$ given the history back from $T-1$. As Transformer models work well when predicting discrete tokens, we transform raw NetFlow traffic into discrete values and use them as model input and target output. The feature representation process is illustrated in more detail in Figure~\ref{fig:embed}.
    }
    \label{fig:pretrain-framework}\postfigcaption
\end{figure}

Given unannotated network traffic data, we use self-supervised learning to exploit intrinsic data correlation and dynamic patterns, yielding general data representations to be utilized for further task-specific fine-tuning.
We view each IP address in the network as a unique node, and the traffic as a time series\footnote{Although traffic can happen at any time, we make abstractions of fixed time granularity such as one minute with aggregated traffic to construct the time series.} pertinent to the node that receives or sends the traffic. We focus on network traffic recorded in the NetFlow \cite{claise2004cisco} format, with the information of IP addresses, timestamps, ports, protocols, packet/byte counts, etc. treated as features into the general ML model.
An overview of our framework is illustrated in Figure~\ref{fig:pretrain-framework}.

\presub\subsection{Pre-training}\postsub

\bfsection{Training Objective}
Similar to next token prediction objectives for NLP foundation models, 
we train the model with the task of predicting the next step of traffic given the history of traffic information in the unlabeled NetFlow data.
But different from language,
the traffic sequence for a node is viewed as a multivariate time series with $f\in\{1 \ldots F\}$ different features.  Our goal is to predict a subset of these features for each of the different nodes in the dataset, a setup that resembles a multi-task learning approach where each node is a task. 
The goal is to provide diverse signals for capturing dependencies and producing general traffic representations for downstream tasks.
At each time step $t$ for each node $v$, let $y^v_{f, t}$ be the $f$-th traffic feature such as number of packets received/sent, timestamp of $t$ such as minute and weekday, number of flows received/sent, etc. The model with learnable parameters $\theta$ is trained by minimizing the following loss
\begin{equation}
    L(\theta) = \frac{1}{VT|\mathcal{F}|}\sum_{v=1}^V \sum_{t=1}^T \sum_{f\in\mathcal{F}} -\log p_{\theta}\left(y^v_{f, t} | \by^v_{<t}, v\right)
\end{equation}
where $V$ is the total number of nodes, $T$ is the total number of time steps, and $\by^v_{ <t}$ denotes all the features on node $v$ prior to time $t$.
The model is tasked to predict a subset of features denoted as $\mathcal{F}\subseteq \{1 \ldots F\}$, for which we maximize corresponding probabilities output by the Transformer decoder model.
Note that for a particular IP node, both the incoming traffic and the outgoing traffic are considered in the traffic features to comprehensively depict the traffic dynamics of the node.
The training objective is general in that it can cover any number of traffic features both in the input side and in the output side, leaving great extensibility for different scenarios.

\begin{figure}
    \centering
    \prefig\includegraphics[width=\linewidth]{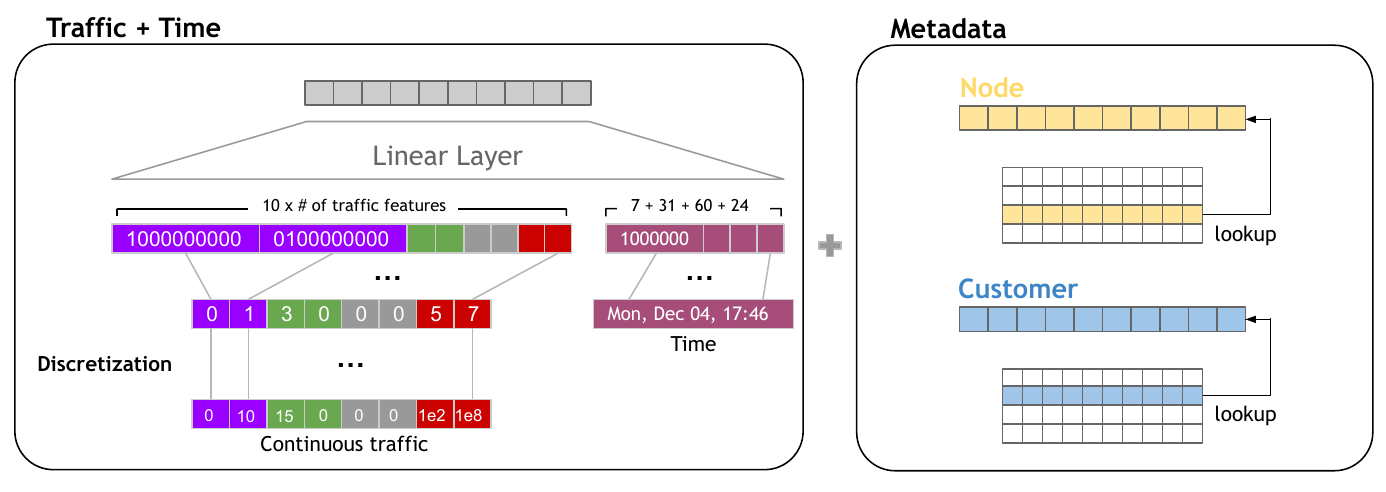}\postfig
    \prefigcaption\caption{Feature representation process of {\name}. We employ two different embedding methods. 
    For inherently continuous features like traffic and time features, we discretize them before transforming them into a consistent fixed-size continuous vector. On the other hand, discrete metadata, such as node ID and customer ID, leverages conventional embedding methods by performing look-ups from trainable embedding tables.
    }
    \label{fig:embed}\postfigcaption
\end{figure}

\bfsection{Feature Representation}
\minlan{Good to start with the classes of network features. and then discuss how we handle each class. Right now, the discussion below is pretty ad hoc with just examples.}\joe{okay, so we can start by dividing features into continuous and categorical? and then go to the handling}\joe{breakdown the flow; start with "we combine the heterogeneous features"; discuss definition of heterogeneous: a) discrete vs. continuous b) different features may affect differently, for the traffic dynamics. refer to table of feature listing. paragraph: for continuous features, xxx. for discrete features, xxx. then finally, how do we combine them.}
Traffic flow features are heterogeneous in different ways. One heterogeneity lies in the intrinsic types of features with two notable classes, one is categorical such as ports and protocols taking discrete values, and another class is continuous such as the number of packets and bytes in NetFlow that can range from zero to arbitrarily large values.
Another heterogeneity comes from varying importance of features, thus affecting traffic dynamics differently. For example, node information can largely indicate traffic patterns on a broad scope, as traffic dynamics could vary drastically between hub nodes and end-user IPs.

We deal with the heterogeneity with unified feature representation pipelines to serve ML models, with principles of minimal customization for features to ensure generalization in the pre-training paradigm. Essentially our feature representation pipeline is based on core properties of all network traffic, which can be adopted uniformly for different traffic features, thus helping the model generalize well when incorporating new features should they arise. 
\minlan{Can you elaborate more on the last sentence? what are the key ideas? what networking domain knowledge do you leverage?}\joe{added a few sentences}
Figure~\ref{fig:embed} gives an overview of the feature representation process.

\begin{algorithm}[tb]
\caption{Continuous Feature Binning Algorithm for Feature Discretization}\label{alg:binning}
\begin{algorithmic}[1]
\Require Original continuous traffic feature time series $\{\bx^v_f\}_{t=1}^{T} \geq 0$, for a particular node $v$ and feature $f$; Number of bins $N$ for the resulting categorical feature.
\Ensure Discretized feature time series $\{\by^v_f\}_{t=1}^{T}$, where each value $y^v_{f, t} \in \{0, 1, 2, \ldots, N-1\}$, taking one out of $N$ categories.

\LeftComment{\textit{Sort all feature values}}
\State Sort $T$ values of $\{\bx^v_f\}_{t=1}^{T}$ in ascending order, denoted with ordering indexes $(t)$ with $\{x^v_{f}\}_{(0)} \leq \{x^v_{f}\}_{(1)} \leq \{x^v_{f}\}_{(2)} \leq \cdots \leq \{x^v_{f}\}_{(T)}$

\LeftComment{\textit{Define the value cutoff points so that each bin has roughly the same number of instances}}
\State Define a sequence of cutoff points $c_0, c_1, \ldots, c_{N-1}$
\State Set $c_0 \gets 0$
\State Find the first value index $k$ such that $\{x^v_{f}\}_{(k)} > 0$, and $\{x^v_{f}\}_{(i)} = 0, \forall i < k$
\State Set $\delta = \mathrm{Round}\left[(T - (k - 1) ) / (N - 1)\right]$
\State Set tentative value cutoff points $c'_{i} = \{x^v_{f}\}_{(k-1 + i * \delta)}, \forall i \in \{1, 2, \ldots, N-1\}$

\LeftComment{\textit{Refine the value cutoff points so that there is no repetition}}
\State Set $c_1 \gets c'_1$
\For {$j \gets 2$ to $N-1$ }
    \If{$c'_j \neq c_{j-1}$}
        \State $c_j \gets c'_j$
    \ElsIf{$c'_j = c_{j-1}$}
        \State Find the first value index $m$ such that $\{x^v_{f}\}_{(m)} > c'_j$
        \State $c_j \gets \{x^v_{f}\}_{(m)}$
        \State $\delta' = \mathrm{Round}\left[(T - (m - 1) ) / (N - j)\right]$, and $c'_{j+1} \gets \{x^v_{f}\}_{(m-1 + \delta')}$
    \EndIf
\EndFor

\LeftComment{\textit{Feature value discretization}}
\State For $t$ such that $x^v_{f, t} = c_0 = 0$, set corresponding $y^v_{f, t}=0$ \Comment{Category 0}
\For{$i \gets 1$ to $N-1$}
\State For $t$ such that $c_{i-1} < x^v_{f, t} \leq c_i $, set corresponding $y^v_{f, t}=i$    \Comment{Category i ($1$ to $N-1$)}
\EndFor

\LeftComment{\textit{The same process could be done for every node $v$ and every continuous feature $f$; however, the continuous values corresponding to the same category class could be different for each $v$ and $h$ combination.}}

\end{algorithmic}
\end{algorithm}

\paragraph{Feature Discretization}
As Transformer models work well when predicting discrete tokens, we 
cast continuous features such as the number of packets into discrete spaces through a \textit{discretization} process. The continuous values of these features are transformed into different levels using a \textit{binning} algorithm.
In particular, for each node $v$ and each feature $f$, we collect all the feature values across all time steps $T$ in the training data, and divide the range of the values into 10 bins. Only zero traffic is cast into the first bin, while non-zero traffic is quantized into the other 9 bins of the same number of instances.
It is essentially finding the cutoff points of equally gapped percentiles of the population of all feature values across time. \joe{double check}
A detailed algorithmic description of the binning procedure is illustrated in Algorithm~\ref{alg:binning}.
This is done separately for each node and each continuous feature; thus the same discrete level category could represent different value ranges for different nodes and features.
Furthermore, the discretization also resolves issues associated with extreme feature values.




\paragraph{Feature Embedding}

After all features are in discrete spaces, they are finally combined into a fixed-size vector through an \textit{embedding} process. This vector is the direct input to the Transformer model, 
encoding all the feature information for further model computation.
We consider features in two groups and embed them differently.
Time and discretized traffic features are first transformed into one-hot encodings and concatenated together. This high-dimensional binary vector is then passed through a linear transformation to form the desired vector length. 
Metadata features such as the node ID and customer ID are encoded with the full embedding length and added to the combined feature embedding vector. These metadata features have a large impact on overall patterns.
Note that we can not enumerate all the node IP values, and our training data cannot cover all the possible values. The pre-trained model will learn useful embeddings of the seen nodes, and we will reuse the embeddings of a most similar node for unseen nodes to generalize our model use case.
More fine-grained IP information, such as the prefixes, are not included here, but could be added to the metadata to learn correlated behaviors of groups of IPs.
Our embedding method is scalable to a variable number of features. Adding more features introduces a trivial amount of extra parameters and does not change the configuration of the backbone transformer model.

\woojeong{}

\minlan{Do you also consider IP prefix? maybe individual IP doesn't have enough traffic but we can learn about the prefix behavior for a group of IPs}\joe{no we haven't considered IP prefix. right now we just enumerate all the existing node IPs in our dataset and give them unique identities, which are represented with different learned embedding vectors to the model}\joe{we can have new embeddings for IP prefixes if we worry about group IPs. /24 2\^24 IPs in the same groups. individual IP may not have enough traffic. different IPs have very different behaviors. attackers can have swithed IPs very frequently. We could list this as futuer work.}
\joe{another concerns (check with Zhiying). supervised learning also formalizes features. what is the differences from engineering the featrues for ML? to clarify to do minimal things for feature representation, to guarantee more generalization. we don't worry about scalablility, as we embedding everything.}
\joe{we focus on netflow, more commonly avaialbe for ISP nowadays. ask Zhiying about it.}

\presub\subsection{Fine-tuning: DDoS Attack Detection}\postsub


The pre-trained model encapsulates general knowledge of network traffic for transfer to tasks with limited annotated labels.
We propose a principled approach to fine-tuning our pre-trained model for downstream tasks, demonstrated through DDoS attack detection.
To demonstrate the power of the pre-trained traffic model for transferring knowledge to various tasks, we also fine-tune the model for the detection of different types of attacks, such as UDP attacks and DNS attacks.
For adapting to new tasks, the fine-tuning process takes only a small amount of annotated traffic data with task-specific labels.
Extensive testing on broader ranges of additional tasks remains a focus for future research.
\minlan{Should highlight that you are fine tuning for different types of DDoS attacks... and what info you need for a new task}\joe{updated.}

Given a time series, we first extract the \textit{traffic hidden representation} from the pre-trained model. This is the internal time series of vectors, which summarizes the traffic dynamics up to the current step. A small classification layer, a simple feedforward neural network, is then added on top of this hidden representation, and fine-tuned with task-specific labels and objectives (see Figure~\ref{fig:pretrain-framework} left). The pre-trained model parameters are kept \textit{frozen} during this process; therefore the fine-tuning phase is lightweight as in the original GPT model \cite{radford2018improving}.

For the application to DDoS attack detection, we target the early detection setup established by Xatu \cite{xu2022xatu}.
Each example $(v, \by, z)$ in the dataset consists of a time series $\by=(\by^v_{1}, \by^v_{ 2}, \ldots, \by^v_{30})$ of 30 minutes with each minute being a time step, and a corresponding label $z\in\{0, 1\}$ marking whether the time series contains an attack or not.
It also contains a manually marked spoof list, so anomaly traffic and normal traffic are also explicitly marked. The goal of this task is to detect the anomaly traffic as much as possible (so the detection will be early), while maintaining a low false positive in marking the normal traffic as attacks.
We adopt the fine-tuning objective as the SAFE loss based on survival analysis \cite{zheng2019safe} tailored to Xatu learning \cite{xu2022xatu}, where the lightweight classification layer outputs non-increasing survival probabilities $s_t$ of the event of an attack not happened yet. By setting up a detection threshold $\tau$, the first time point $t$ when $s_t<\tau$ will be detected as the attack time.
\joe{Xatu evaluation? -> expeirmental setup. SAFE loss details and math? maybe no space.}

\presec\section{Dataset Construction}\postsec

\begin{table}[]
    \centering
    \pretablecaption\caption{Statistics of our NetFlow Data for traffic pre-training.}
\label{table:netflow}\posttablecaption
    \pretable\begin{tabular}{ccc}
    \toprule
    & Raw & Filtered
    \\\midrule

    Customer & 87 & 80 \\
    Node (IP) & 6,975 & 520 \\
    Training Size & 1,297,350 & 96,720\\
    Validation Size & 146,475 & 10,920 \\
    \bottomrule
    \end{tabular}\posttable

\end{table}

\begin{table}[]
    \centering
    \pretablecaption\caption{Statistics of {\xatu} Data for DDoS attack detection.}
\label{table:xatu}\posttablecaption
    \pretable\begin{tabular}{ccc|cccc}
    \toprule
    & Nodes & Examples & Total Attack & DNS & UDP & NTP \\
    \midrule
    Train & 377 & 1606 & 803 & 255 & 412 & 92 \\
    Val & 45 & 360 & 180 & 13 & 139 & 25 \\
    Test & 135 & 406 & 203 & 106 & 63 & 21 \\
    \bottomrule
    \end{tabular}\posttable

\end{table}

\subsection{NetFlow Data Serialization}
We employ a pre-training dataset obtained from a large ISP, which is a comprehensive collection of sampled NetFlow records captured within the ISP's network over a span of three months, accompanied by 16K attack alerts sourced from a widely utilized commercial defense system. Specifically, we extract the inbound and outbound traffic pertaining to each target IP address, employing a temporal resolution of one minute. The dataset consists of time series total length of 132,480 minutes. We partition this dataset along the time axis into distinct sets for training, validation, and testing. The training set consists of 95,232 minutes, while the validation and testing sets span 10,752 minutes and 26,496 minutes, respectively. We further divide each dataset into discrete time series units of 512 minutes for pre-training.

\subsection{NetFlow Data Filtering}
We identified a certain customer that generates a substantial amount of traffic and attacks during a specific time frame. To ensure the generalizability of our model to other customers, this dominant and any associated customers are excluded from our analysis. Collectively, these customers accounted for approximately 92\% of the total number of customer IP addresses. The resulting statistics of the filtered NetFlow data are summarized in Table~\ref{table:netflow}.

We adopted the alerts from the DDoS defense of our NetFlow provider as ground-truth labels and filtered out the following alerts: (1) potential false positive alerts reported by the DDoS defense; (2) mismatched alerts with reported packets and bytes less than 95\% of packets and bytes observed in the NetFlow. We focus on the most prevalent three attack types after filtering: DNS Amplification attacks, UDP attacks, and NTP Amplification attacks. 

\bfsection{Traffic Features}
\begin{table}[]
\pretablecaption\caption{List of traffic features.}
\label{table:traffic-features}\posttablecaption
\pretable\begin{tabular}{lccc|c}
\toprule
\multicolumn{2}{c}{Feature type} & In/Out & Traffic type & \#  \\
\midrule
\multicolumn{2}{c}{Volume} & both & pkt, byt, flow & 6 \\
Protocol & {ICMP, TCP, UDP, other} & both & {pkt, byt, flow} & 24 \\
Port & {0, 53, 80, 123, 443, well-known, registered, private} & both & {pkt, byt, flow} & 42\\
TCP flag & {0, 16, 24, 2, 17, 18, 4, 25} & out & flow & 8 \\
\bottomrule
\end{tabular}\posttable

\end{table}
Table~\ref{table:traffic-features} presents a list of the traffic features employed during the pre-training phase. \textit{Volume} features refer to the aggregated traffic flow from all protocols, TCP flags, and ports. Our approach also considers more fine-grained features by grouping incoming or outgoing traffic by protocol, TCP flags, and ports. In terms of ports, \textit{well-known} refers to ports ranging from 0 to 1023, excluding those previously specified. Ports in the range of 1024 to 49151 are categorized as \textit{registered} while ports spanning 49152 to 65535 are designated as \textit{private}. Incoming traffic is aggregated based on the destination port, while outgoing traffic is aggregated according to the source port. Furthermore, we segment the traffic based on the eight most common TCP flags, in the order of \textit{URG, ACK, PSH, RST, SYN, FIN}. Overall, our approach leverages a total of 86 traffic features. Our embedding method alleviates the necessity of manual feature selection, as it has the capacity to encode any desired number of features.

\subsection{Downstream Task: Early DDoS Attack Detection}
\minlan{Throughout the paper, rather than Xatu data, call it Tata data (same as how Xatu describes the data set) and then say the label and/or baseline is from the Xatu paper.}
To form the data for downstream fine-tuning, we select an equal number of attack and non-attack time series for each customer IP address based on ground-truth alerts. For each alert, we generate the attack time series for this alert and randomly select a non-attack time series from the same customer IP address within the same minute range of data split. For each time series, we extract time series of 86 features in the 512 minutes. We also identify the onset of the anomaly for the attack time series by looking for the sudden, sustained increase in the matching traffic, using CUSUM algorithm~\cite{carl2006denial, grigg2003use}. The time difference between this anomaly onset and attack detection indicates the timeliness of a detection approach. Table~\ref{table:xatu} shows the statistics of this data, named as {\xatu}. 
 

\notes{
\subsection{Data Analysis}

Distribution, visualization, etc.

e.g. show a few different patterns of different nodes, some are more static, some are periodic, etc.

(no space now)
}


\presec\section{Experimental Setup}
\label{sec:experiments}\postsec

\begin{figure}[t]
     \centering
     \begin{minipage}{0.8\linewidth}
    \includegraphics[width=\linewidth]{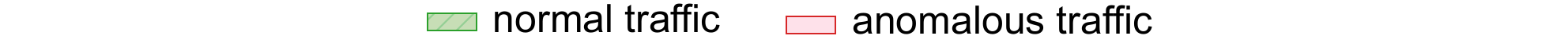}
    \end{minipage}
    \begin{minipage}{0.45\linewidth}
    \prefig\includegraphics[width=\linewidth]{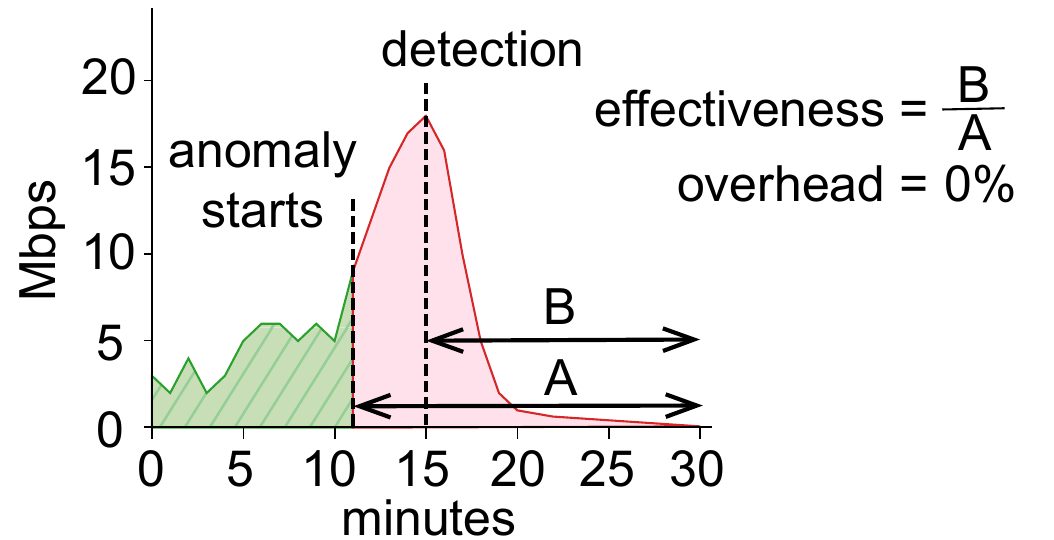}\postfig
    \subcaption{Late detection}
    \end{minipage}
    \begin{minipage}{0.45\linewidth}
    \prefig\includegraphics[width=\linewidth]{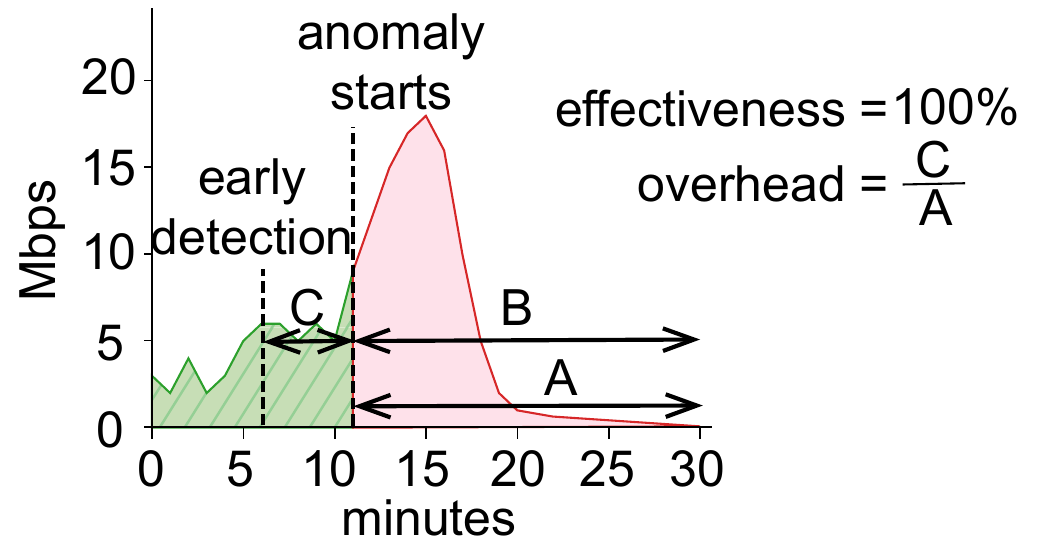}\postfig
    \subcaption{Early detection}
    \end{minipage}
     \prefigcaption
     \caption{An example UDP attack describing the effectiveness and overhead metrics. A: anomalous traffic; B: anomalous traffic detected; C: extraneous traffic detected. Given a detection time, effectiveness measures the anomalous traffic that is detected, and overhead measures the normal traffic that is erroneously classified as attacks. (a) shows an example of the late detection case, and (b) shows an example of an early detection case.} \postfigcaption
     \label{fig:metrics}
 \end{figure}

\notes{

brief intro of Xatu evaluation metric.

(or defer evaluation metrics close to results?

}

\subsection{Model \& Data Configuration}
We pre-train a Transformer decoder model with 4 layers, 4 attention heads, embedding and pre-trained representation size 128, and intermediate feedforward dimension 512, resulting 1.9 M total number of parameters.
The model learns to predict all 86 traffic features in $\mathcal{F}$ during training for every minute, and history traffic if fed with $T=512$ minutes.
For fine-tuning, instead of using a large set of labels, we only employ the {\xatu} validation set.
The lightweight classification has one hidden layer of size 512, totaling 67K parameters.
Our objective is to accurately predict DDoS attacks during the test time span, which encompasses a total of 203 attacks.

\subsection{Evaluation}
For pre-training, we evaluate the goodness of next step traffic prediction with \textit{perplexity} (PPL) \cite{bengio2000neural} scores,
defined as $\mathrm{PPL}=\exp(L(\theta))\geq 1$.
A low value of perplexity indicates the model probability distribution predicts the ground truth feature well.
As all the features are discretized in the model, we also measure the \textit{accuracy} of traffic predictions. These values are dependent on the discretization used, and are mainly useful for measuring training progress.

\woojeong{detailed description on evaluation} For the downstream task of DDoS attack detection, we evaluate based on the \textit{effectiveness} with bounded \textit{overhead} metrics from Xatu \cite{xu2022xatu}.
Given an attack detection time from a model, as illustrated in Figure~\ref{fig:metrics}, effectiveness measures how much percent of anomalous traffic is mitigated with this detection, and overhead measures how much normal traffic is erroneously labeled with attacks. To tackle the trade-off between effectiveness and overhead, we select an overhead and maximize effectiveness while bounding a certain percentage of customers within the selected overhead. This percentage is overhead bound. The effectiveness and overhead metrics are conceptually correlated with the true positive rate (TPR) and false positive rate (FPR) in the traditional binary classification setup, which we also report for detection accuracy.

\subsection{Model Implementation}
We adopt a Transformer decoder model with 4 layers, 4 attention heads, an embedding and hidden representation size of 128, and an intermediate feed-forward dimension of 512, with the total number of parameters being 1.9 million. The model is trained for 100 epochs at maximum, using a learning rate of 0.0005, and Adam optimizer \cite{kingma2014adam}.
Training is on mini-batches, where each batch contains random samples of traffic time series for IP nodes (could be different nodes) of up to 512 minutes.
We build our Transformer models with the Fairseq \cite{ott2019fairseq} implementations using PyTorch \cite{paszke2019pytorch} deep learning library.




\presec\section{Main Results}
\label{sec:results}\postsec

\presub\subsection{Traffic Predictive Modeling}\postsub

\begin{table}[t]
    \centering
    \pretablecaption\caption{Pre-training results of next traffic prediction.\joe{average PPL: 1.52 for baseline. these are for the validation set with no_ishan customers.}
}
\label{table:pretraining}\postfigcaption
    \pretable\begin{tabular}{cccc}
    \toprule
    & Loss $\downarrow$ & Perplexity $\downarrow$ & Accuracy $\uparrow$
    \\\midrule
    Pre-trained & 0.19 & 1.20 & 0.94 \\
    Baseline & 0.21 & 1.23 & 0.93 \\
    \bottomrule
    \end{tabular}\posttable

\end{table}

\notes{
\vspace{0.2in}

\textbf{Results planning:
}

\begin{itemize}
    \item Main table for the PPL, with different models and baselines (bi-gram model); both for training data and for validation data; averaged PPL, over all nodes, all features
    \item detailed PPLs of nodes and features, e.g. a heatmap
    \item One qualitative graph: traffic tracking of one feature, e.g. pkt or byt: true traffic, predicted traffic (recovered from discretization, use the mean/median of the bin)
\end{itemize}


(TBD) PPL of each feature, MSE on the recovered signal, visualization.

\vspace{0.2in}
}

We first measure how well the pre-trained model captures traffic dynamics directly by predicting the next step traffic, which is a check on the pre-training task.
We compare with a baseline of a statistical model that estimates the probability of a feature value based on the immediate previous value, similar to the bigram language model \cite{ney1991smoothing}. This baseline is strong as most features remain constant each time step.

Table~\ref{table:pretraining} presents a sanity check of traffic prediction performance on validation data. The metric scores are averaged over all nodes and features.
We can see that the generative pre-trained model performs better than the baseline, indicating the dependency on history traffic information is being learned to capture the traffic patterns.

\notes{
We also show an example of how the predicted traffic tracks the real traffic in Figure~\ref{fig:pretrain-result}. \joe{a qualitive example here if we have space --> Seems not.}

}

\presub\subsection{Early DDoS Attack Detection}\postsub

\begin{table*}[t]
    \centering
    \pretablecaption\caption{Comparison between our model and the baselines on all types of attacks with overhead bound of 80\%. Per attack type, the best numbers of each row are indicated in bold.}
\label{table:attack-all}\posttablecaption
    \pretable\begin{tabular}{lcccccc}
    \toprule
    & {\name} & \multicolumn{2}{c}{Transformer} & \multicolumn{2}{c}{Multiscale-LSTM}
    \\\midrule
    Training samples & 360 & 360 & 1606 & 360 & 1606
    \\\midrule
    Name & \texttt{\name} & \texttt{TF} & \texttt{TF\_MoreData} & \texttt{MS-LSTM} & \texttt{MS-LSTM\_MoreData} \\
    \midrule
    Effectiveness $\uparrow$ & \textbf{100.00} & 81.36 & \textbf{100.0} & 99.27 & \textbf{100.00} \\
    Overhead $\downarrow$ & 0.06 & \textbf{0.03} & 0.09 & 0.07 & 0.08 \\
    FPR $\downarrow$ & \textbf{3.45} & 6.40 & 6.40 & 12.81 & 8.87\\
    FNR $\downarrow$ & 5.42 & 15.76 & 3.45 & \textbf{0.00} & 0.49 \\
    F1 score $\uparrow$ & \textbf{0.96} & 0.88 & 0.95 & 0.94 & \textbf{0.96} \\
    \bottomrule
    \end{tabular}\posttable
\end{table*}

\paragraph{Baseline}
 We compare our method to two baseline setups: models trained with the same amount of labeled data, and models trained with additional labeled data. For the latter, the models were trained on the {\xatu} train split, which contains five times more training examples. Regarding the model configurations, we employed two baseline models. The first one is a Transformer model without pre-training. To demonstrate the impact of pre-training, we randomly initialized a Transformer model with the same architecture as our model. The second baseline model is a Multiscale-LSTM~\cite{xu2022xatu}, which consists of multiple LSTMs in various downsampled timescales. These two baseline models see the same length of history (512), the number of traffic features (86), and training objective (SAFE loss), but unlike the pretrained model are trained only on DDoS data.

\paragraph{Main Results}
The main results for all types of attacks are presented in Table~\ref{table:attack-all}. 
When compared to the Transformer without pre-training and Multiscale-LSTM trained with the same number of examples, our model demonstrates the highest effectiveness while maintaining a comparable overhead. The Transformer model exhibits a higher FNR, indicating its inability to detect actual attacks accurately. 
For predicting potential attacks, the FPR becomes more crucial. LSTM models tend to overestimate false positives, identifying 12.81\% of false positive attacks. In contrast, our method showcases a balanced performance across all metrics. 


\minlan{You probably need to redescribe the metrics in Xatu clearly here as well. Assume the readers havn't read the Xatu paper, all the info should be self-contained.}

The pretrained approach also exhibits comparable performance to models trained with a significantly larger number of annotated labels. Even when trained with only 22\% of the labeled data, our model outperforms the Transformer model on F1. Considering the challenges of acquiring a large number of labels in practical scenarios, the semi-supervised pre-training stage proves to be highly advantageous. It enables transfer learning of pretrained models to downstream tasks, even with the limited number of ground truth labels.

\begin{figure}
    \centering
    \prefig\includegraphics[width=0.9\linewidth]{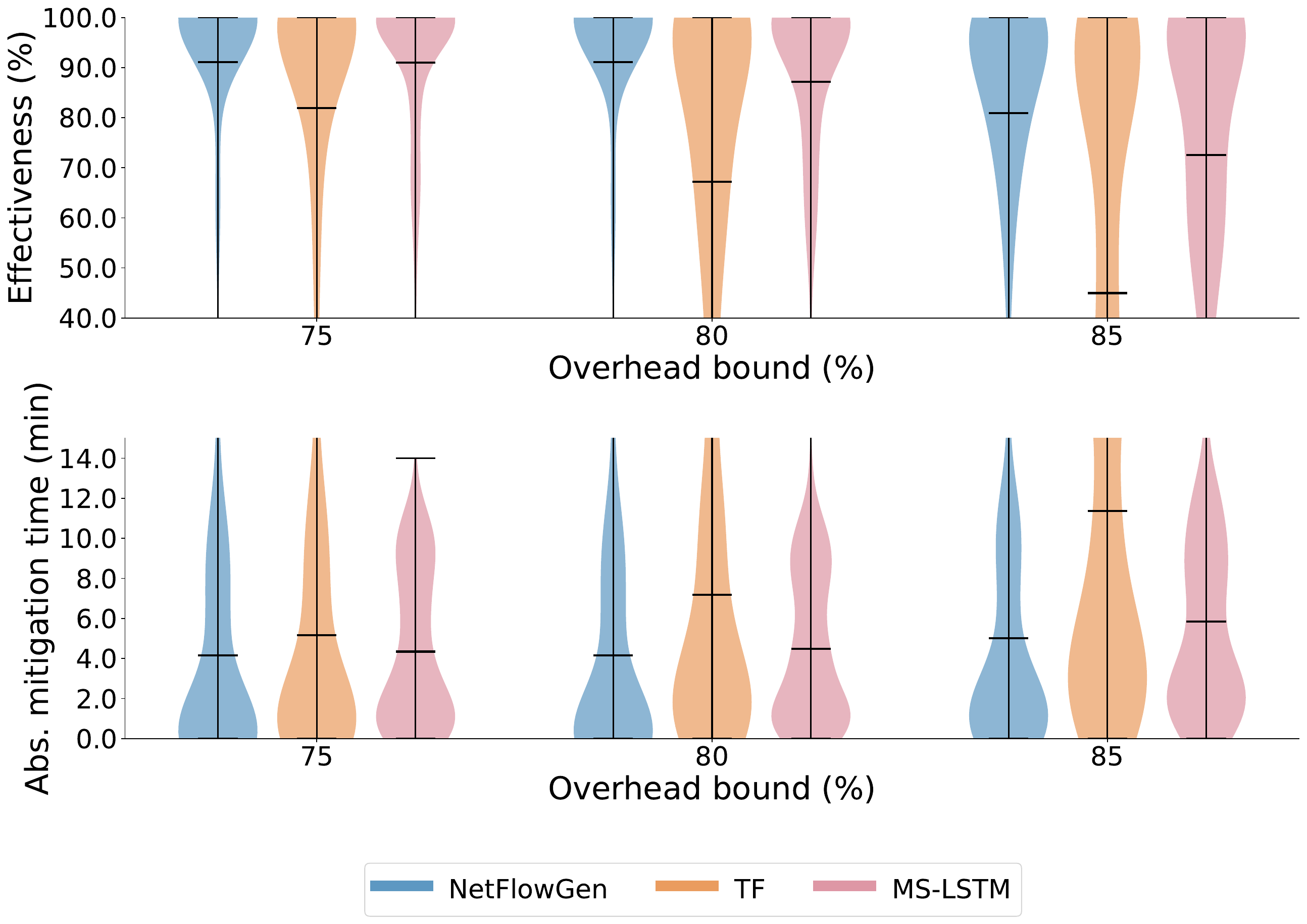}\postfig
    \prefigcaption\caption{
    Effectiveness and absolute mitigation time under three different overhead bounds
    }
    \label{fig:main}\postfigcaption
\end{figure}

Figure~\ref{fig:main} presents a comprehensive comparison of different models under varying overhead bounds, by depicting the violin plots of effectiveness and absolute mitigation time distributions.
Our model demonstrates robustness across all three overhead bounds and consistently outperforms the baselines, seen by our more concentrated distributions on higher effectiveness values, particularly under more conservative overhead constraints.
Additionally, our model excels in predicting attack times, achieving the lowest absolute mitigation time, which reflects how early the model detects an attack.
Moreover, our method predicts attack times with high accuracy and minimal variance, highlighting its precision and reliability.

\woojeong{TODO: make it 60-100, check what error bars are for, address why error bars are big (since labeling is hard, we can't )}


\paragraph{Mitigation time} In terms of predicting the time of attack, our model excels compared to the baselines. Figure~\ref{fig:main} illustrates the absolute mitigation time, indicating how accurately a model detects an attack. The optimal scenario is a mitigation time of zero. Our method demonstrates the closest mitigation time to zero, with the smallest variance.

\paragraph{Detection per attack type}
The advantage of a pretrained network traffic prediction model lies in its ability to transfer knowledge to various networking tasks through fine-tuning. We evaluate the performance of {\name} across different attack types, including DNS, UDP, and NTP attacks. 
Fine-tuning is conducted using the {\xatu} validation set with the corresponding attack type. For a fair comparison, the baseline models are trained using an equal number of training samples. 
Table~\ref{table:attack-pertype} provides an overview of the effectiveness of our method in detecting various types of attacks.\minlan{Maybe elaborate more on the takeaways from the table. Compare across attack types, cite specific numbers etc...} Our pretrained model exhibits better generalization capabilities across all attack types in terms of effectiveness, reducing the reliance on task-specific labels. Specifically, it is worth noting that DNS attacks are relatively rare during the training phase, with only 13 instances observed in Table~\ref{table:xatu}, yet our pretrained model predicts 203 test attacks at the nearly exact time.
\minlan{Give a few more description on where the benefits come from and why it's a good result}

\begin{table*}[]
    \centering
    \pretablecaption\caption{Comparison between our model and the baselines on each of DNS, UDP, and NTP attack detection tasks. The best numbers of each row are indicated in bold.}
\label{table:attack-pertype}\posttablecaption
    \resizebox{\columnwidth}{!}{
    \pretable\begin{tabular}{lccccccccc}
    \toprule
    Attack Type & \multicolumn{3}{c}{DNS} & \multicolumn{3}{c}{UDP} & \multicolumn{3}{c}{NTP} \\
    \cmidrule(lr){2-4} \cmidrule(lr){5-7} \cmidrule(lr){8-10}
    Model & \name & Transformer & LSTM & \name & Transformer & LSTM & \name & Transformer & LSTM\\
    \midrule
    Effectiveness $\uparrow$ & \textbf{100.00} & \textbf{100.00} & 63.83 & \textbf{95.96} & 91.31 & 53.14 & \textbf{99.87} & 98.81 & 93.78\\
    Overhead $\downarrow$ & \textbf{0.00} & 0.06 & 0.05 & \textbf{0.00} & 0.01 & 0.06 & \textbf{0.00} & \textbf{0.00} & 0.09 \\
    FPR $\downarrow$ & \textbf{2.83} & 13.21 & 16.04 & 9.52 & \textbf{3.17} & 4.76 & \textbf{0.00} & \textbf{0.00} & 14.29\\
    FNR $\downarrow$ & \textbf{0.00} & 0.94 & 0.94 & 26.98 & 33.33 & 6.35 & 9.52 & \textbf{4.76} & 14.29\\
    F1 score $\uparrow$ & \textbf{0.99} & 0.93 & 0.92 & 0.80 & 0.79 & \textbf{0.94} & 0.95 & \textbf{0.98} & 0.86\\
    \bottomrule
    \end{tabular}\posttable
}

\end{table*}

\section{Ablation Study}
In this section, we aim to delve deeper into the effectiveness of \name~by systematically analyzing its performance across diverse scenarios.
Firstly, we scrutinize its ability to generalize to unseen nodes, assessing its adaptability to different traffic patterns. Next, we explore the impact of diverse traffic features on the model's predictive capabilities, evaluating how different types of traffic inputs affect performance. Lastly, we analyze the effect of varying Transformer model sizes on the model's pre-training performance, aiming to decide the optimal size for our task.

\notes{
\subsubsection{Qualitative Analysis}
Detailed analysis of examples, false positive, false negative, etc. \woojeong{add a false positive example if we have space}
}

\minlan{give some preamble on what ablation study you did and what questions you are trying to answer here}




    










\subsection{Generalization to Unseen Nodes}

\begin{table}[t]
    \centering
    \pretablecaption\caption{Nearest-neighbor search results of unseen nodes of the validation set.}
\label{table:disc}\postfigcaption
    \pretable\begin{tabular}{ccc}
    \toprule
    & Node 3679 & Node 6677 \\
    \midrule
    Total attacks received & 4 & 155 \\
    The number of mapped unseen nodes & 205 & 15\\
    \bottomrule
    \end{tabular}\posttable
\end{table}

We consider applying {\name} to nodes unseen during pre-training, especially those that were not subjected to any attacks and have different traffic patterns. We select 432 {\xatu} examples each from the validation and test time spans. Since these nodes were not seen during the pre-training process, we use a nearest-neighbors search to find the most similar training node and apply its discretization splits.
Since traffic flow features are high-dimensional ($V \times T \times |\mathcal{F}|$) and heterogeneous depending on nodes and feature types, we first summarize each node's traffic features through the following process. The features of each node are standardized and reduced to minimum, maximum, and 25th, 50th, and 75th percentile statistics, along the time axis. Next, we average features, representing each node as a vector of the five statistics. 
The closest match is found by computing the L1 distance between an unseen node and all training nodes. Through this nearest-neighbors approach, every unseen node is mapped to the most similar training node. As shown in Table~\ref{table:disc}, non-attack nodes are mapped to a very small number of nodes that have monotonic traffic patterns.
\minlan{what does node 3679, 6677 mean? again, mention specific numbers in text and highlight more on your design that enables this result.}


The classification results for unseen node traffics are presented in Table~\ref{table:nonattack}. Given the relatively low magnitudes of these traffics, {\name} can easily identify the absence of any actual attacks. It accurately classifies all 432 test examples as non-attack instances.

\begin{table}[]
\centering
\pretablecaption\caption{Confusion matrix of {\name} on unseen nodes.}
\label{table:nonattack}\posttablecaption
\pretable\begin{tabular}{@{}c cccc|ccccc@{}}
\toprule
\multicolumn{2}{c}{} & \multicolumn{2}{c}{Predicted} &  & \multicolumn{2}{c}{} & \multicolumn{2}{c}{Predicted} &  \\ \cmidrule(lr){3-4} \cmidrule(lr){8-9}
\multicolumn{2}{c}{\multirow{-2}{*}{\textbf{Validation set}}} & True & False & \multirow{-2}{*}{Total} & \multicolumn{2}{c}{\multirow{-2}{*}{\textbf{Test set}}} & True & False & \multirow{-2}{*}{Total} \\ \midrule
\multicolumn{1}{c}{} & True & 0 & 0 & 0 & \multicolumn{1}{c}{} & True & 0 & 0 & 0 \\ \cmidrule(lr){2-5} \cmidrule(l){7-10} 
\multicolumn{1}{c}{\multirow{-2}{*}{Actual}} & False & 2 & 430 & 432 & \multicolumn{1}{c}{\multirow{-2}{*}{Actual}} & False & 0 & 432 & 432 \\ \bottomrule
\end{tabular}\posttable

\end{table}

\subsection{Traffic Feature Diversity}

\begin{table*}[]
    \centering
    \pretablecaption\caption{Comparison between different sizes of pre-training data with overhead bound of 80\%. The best numbers of each column are indicated in bold.} \label{table:light}\posttablecaption
    \resizebox{\columnwidth}{!}{
    \pretable\begin{tabular}{ccccccc}
    \toprule
    & Traffic feats & Effectiveness $\uparrow$ & Overhead $\downarrow$ & FPR $\downarrow$ & FNR $\downarrow$ & F1 score $\uparrow$ \\
    \midrule
    \texttt{\name} & 86 & \textbf{100.00} & 0.06 & \textbf{3.45} & 5.42 & \textbf{0.96} \\
    \texttt{\name\_Light} & 6 & 99.96 & 0.06 & 12.81 & \textbf{2.96} & 0.92\\
    \bottomrule
    \end{tabular}\posttable
}
\end{table*}

In order to show the effect of using more diverse traffic features during pre-training, we present the result of using only a subset of features.
\name~aggregates 86 network traffic features in total by protocol, TCP flags, and the port associated with incoming or outgoing traffic. Table~\ref{table:traffic-features} is a full list of the traffic features used by \name. 
We built \name\_Light using only the "Volume" features of Table~\ref{table:traffic-features} without the fine-grained ones. Since our embedding method is scalable to a variable number of features, the configuration of the backbone transformer model is the same for \name~and \name\_Light.
Table~\ref{table:light} demonstrates that \name~trained on full 86 features outperforms \name\_Light, especially on FPR and F1 score. Still, \name\_Light shows comparable performance to the baseline models in Table~\ref{table:attack-all}, showing the effectiveness of the pre-training approach.

\subsection{Transformer Model Size}

\begin{table*}[]
    \centering
    \pretablecaption\caption{Comparison between different sizes of the transformer model. "Model Config." column denotes \{\# of layers\}-\{\# of heads\}-\{hidden size\}-\{intermediate hidden size\}. LR is the learning rate, Dropout is the random dropout probability during training. The best numbers of each column are indicated in bold.} \label{table:model_sizes}\posttablecaption
    \resizebox{\columnwidth}{!}{
    \pretable\begin{tabular}{cccccccc}
    \toprule
    \multicolumn{1}{l}{Model Config.} & \multicolumn{1}{l}{\# of Params.} & \multicolumn{1}{l}{LR} & \multicolumn{1}{l}{Dropout} & \multicolumn{1}{l}{Best Val Loss} & \multicolumn{1}{l}{Best Val Acc.} & \multicolumn{1}{l}{Best Val Ppl.} & \multicolumn{1}{l}{Best Val F1} \\
    \midrule
    4-4-128-512 & 1.9M & 0.0005 & 0.3 & 0.186 & \textbf{0.935} & 1.204 & 0.983 \\
    \multirow{4}{*}{4-4-256-1024} & \multirow{4}{*}{5.4M} & \multirow{2}{*}{0.0001} & 0.1 & 0.184 & 0.936 & 1.202 & 0.983 \\
     &  &  & 0.3 & 0.186 & \textbf{0.935} & 1.204 & 0.983 \\
     &  & \multirow{2}{*}{0.0005} & 0.1 & 0.182 & 0.936 & \textbf{1.199} & 0.983 \\
     &  &  & 0.3 & 0.183 & 0.936 & 1.200 & 0.983 \\
    4-4-512-1024 & 12M & 0.0005 & 0.3 & \textbf{0.181} & 0.936 & \textbf{1.199} & 0.983 \\
    4-4-512-2048 & 25M & 0.0005 & 0.3 & 0.182 & 0.936 & 1.200 & 0.983 \\
    \bottomrule
    \end{tabular}\posttable
    }
\end{table*}

We also experiment with various sizes and configurations of the pre-trained transformer model and present the pre-training results in Table~\ref{table:model_sizes}. We find that our model is agnostic to the sizes we tested. The smallest model, with 4 decoder layers, 4 attention heads, 128 model hidden size, and 512 intermediate hidden size, is used for further fine-tuning experiments as it's more efficient during training and inference bu already achieves good performance.
Note that the loss patterns are similar to the recent observation in NLP that bigger models are more powerful in terms of data efficiency \cite{kaplan2020scaling}, which lead to lower loss eventually with a lot of data.
Compared to the loss, the validation accuracy takes a cruder set of values, which is less sensitive to the model sizes. This is likely related to the traffic data quality in our pre-training, which has lower noise as a result of our data filtering. The relationship between model size and data quantity and quality remains an interesting direction for future works.
\notes{
the optimal model size is coupled with the data dynamics
according to scaling laws, bigger models are more powerful in that they are more data-efficient and can eventually reach better loss with a lot of data, but inefficient
small model size is an effect of the data quality or our way of filtering the data.
the optimal size of the model relevant to data is in question and can be put to future works
}

\presec\section{Related Work}\postsec

Statistical methods have been widely used to understand traffic dynamics in networks and are commonly based on probabilistic models to track network behavior and detect changes over hard thresholds. The entropy of traffic features, as a basic but prevalent metric, provides fine-grained insights for anomaly detection~\cite{feinstein2003statistical, nychis2008empirical, berezinski2015entropy}. Wavelet analysis reveals wavelet components through wavelet transformation and captures anomalies from the changes in these components~\cite{hamdi2007detecting, callegari2011combining}. Principal component analysis~(PCA), a dimensionality reduction approach, efficiently separates traffic measurements into normal subspace for principal components and anomalous subspaces for the rest~\cite{lakhina2004diagnosing, ring2019flow, kanda2013admire}. Other statistical methods employ covariance matrices to track the alterations between normal traffic and flooding attacks~\cite{yeung2007covariance}, hidden semi-Markov model to depict the spatial-temporal characteristic of the wake-up packet generation process~\cite{bang2017anomaly}, CUSUM algorithms to identify unusual deviations in traffic patterns from the norm in real-time with minimal memory expenses~\cite{van2015real}.
In contrast to statistical methods that rely on careful design and domain intuition, we employ machine learning, particularly deep learning, with end-to-end training that is data-driven and highly scalable.

The widespread emergence of machine learning approaches offers fresh opportunities to comprehend network traffic dynamics. Classical learning-based approaches, including clustering~\cite{karami2015fuzzy, qin2015ddos, roudiere2018evaluating}, naive Bayesian~\cite{singh2016distributed, swarnkar2016ocpad}, Support Vector Machine~(SVM)~\cite{meti2017detection, kabir2018novel}, decision tree~\cite{jacobs2022ai}, and random forest~\cite{idhammad2018detection}, are broadly adopted to identify and classify network traffic. These approaches can be combined together for better performance. For example, a two-step detection approach in IXP scrubber~\cite{wichtlhuber2022ixp} first tags each flow with association rule mining and then classifies per-target IP profiles aggregated from flows with ML classifier.
Apart from these classical approaches, deep neural networks have also been designed to handle even more complex tasks and capture intricate patterns in network traffic in recent years. Doshi et al.~\cite{doshi2018machine} uses neural networks on IoT-specific network behaviors to detect anomalies generated from a local network. Kitsune~\cite{mirsky2018kitsune} tracks features in network channels and detects network intrusion with an ensemble of auto-encoders. Instead of directly classifying traffic patterns, deep neural networks are also capable of learning traffic representatives as an intermediary step for downstream endeavors of various purposes. Meng et al.~\cite{meng2023netgpt} provides general encoding for network traffic and optimizes the adaptation effect of the model to diversified tasks; ET-BERT~\cite{lin2022bert} pre-trains deep contextualized datagram-level representation from large-scale unlabeled traffic data.
Both of these approaches adopt the pre-training approach with unlabeled data, but they are tailored to specific networking features, which limits their generalization potential.

Despite the abundance of networking traffic data, building a unified pre-trained foundation model \cite{bommasani2021opportunities} for networking presents unique challenges. Networking data is multivariate and highly heterogeneous, comprising features such as IP addresses, ports, transport protocols, timestamps, packets, and bytes, which vary in type (continuous vs. categorical) and scale. This diversity requires an effective pre-training objective capable of capturing broad network dynamics while balancing feature importance and handling data heterogeneity.
Additionally, curating well-balanced datasets for general-purpose pre-training is difficult. Moreover, for benchmarking tasks, ground truths, such as true attack labels in network security, are often unavailable, expert labeling is costly, and privacy concerns further constrain data collection. Prior works \cite{dietmuller2022new, le2022rethinking} have focused on theoretical discussions or small-scale simulations. In contrast, our framework addresses these challenges with real-world, large-scale network traffic of diverse features, using data collected in the wild for training and evaluation to ensure relevance to critical network applications.

\joe{flow-based traffic vs. packet information on micro level}

\label{sec:related-work}

\section{Discussion and Improvement Potentials}
\label{sec:discussion}

We present a practical application of the self-supervised machine learning framework on network dynamics modeling, by pre-training a foundation model based on Transformer decoders with large amount of traffic data, and fine-tuning on specific downstream networking tasks. The self-supervised foundation models \cite{bommasani2021opportunities} have seen great successes in other fields such as NLP \cite{radford2018improving, radford2019language, brown2020language}, but there exist unique challenges for networking data due to its disparity in data and applications.
Despite our effort in bringing generative model pre-training techniques onto network traffic data with both domain customization and model unification, there are still open questions and limitations around our method that are worth noting, which could also provide potential for further improvements.




\bfsectionn{How to best handle the traffic features?}
Although both can be naturally formulated as sequences or time series, networking traffic data is very different from text data in NLP, where Transformer-based foundation models are widely applied. First, text sequences have regular discrete indexes as a time series, whereas traffic data can occur at any time with irregular time increments. Second, text data are discrete tokens with a finite number of values coming from a well-defined vocabulary, whereas traffic data are continuous and can take arbitrary values. As Transformers \cite{vaswani2017attention} are initially designed and also best known for processing text data, we customize the network traffic feature representation pipeline to make the features closer to their NLP counterpart. This includes aggregating traffic under one-minute intervals to make regular time series, and discretizing traffic values into categorical classes. However, our procedures also pose limitations on the full potential of utilizing these features. For example, traffic aggregation may lose details of when and how the traffic occurs, and feature discretization is a lossy process that can not be easily recovered. There might exist better ways of incorporating more details of the networking traffic that can benefit downstream applications that are more sensitive to this detailed feature information.

\bfsectionn{How to incorporate node interactions?}
In our pre-training formulation, the traffic time series is constructed per node or IP address. For each node, we record both the incoming and outgoing traffic, and the {\name} model is trained on all the time series for all available nodes. However, no explicit IP node interactions are being learned during this process. For example, the model does not have the knowledge of which node sends traffic to which other nodes. This information could be critical in applications, such as P2P traffic identification \cite{iliofotou2007network} and reverse protocol analysis \cite{zhang2022dual}, that require not only the traffic volumes, but also the IP addresses that send or receive the traffic. In other words, the network topology is not incorporated during the learning process. Potential improvements could directly encode node information when formulating the traffic features, or adopt a different pre-training model architecture such as graph neural networks (GNNs) \cite{wu2020comprehensive, li2019hierarchy, hu2019strategies} which take into consideration the graph topologies. This is likely to depend on different designs of data formulation and training objectives.

\bfsectionn{How big should a networking foundation model be?}
Although we curate a network traffic pre-training dataset of about one million minutes from a real ISP that is considered large-scale compared to the data used in previous works, by the standard of foundation models in other fields such as NLP, the size of our pre-trained models are not large in the number of parameters. For example, the state-of-the-art large language models (LLMs) contain billions of parameters \cite{brown2020language, chowdhery2022palm}, and it is shown that the model performance in downstream applications is increasing with the size of the model \cite{brown2020language, kaplan2020scaling, ding2020statistical, hoffmann2022training}. Does the same trend hold for networking data, and what size of the model should be considered the LLM for network traffic foundation model with an exceptional level of overall understanding of the network dynamics? This would likely call for more actions in networking research on data collection, downstream benchmark constructions, and pre-trained model scaling. Although networking data is abundant, large-scale data collection could pose privacy concerns. Furthermore, both data quantity and data quality would play important roles in building stronger networking foundation models, as well as the intrinsic data complexities.

\presec\section{Conclusion}
\label{sec:conclusion}\postsec

We envision a general framework of pre-training an ML model to capture network traffic dynamics for adaptation to downstream tasks, utilizing large amount of raw networking data and circumventing the expensive labeling issues.
As an early attempt, we propose {\name} with designs to represent heterogeneous NetFlow features in a unified space for model processing, and adopt  Transformer decoder model with a multivariate generative pre-training task. We construct a realistic large-scale dataset for the pre-training and test the model adaptation on a downstream DDoS attack detection task. Experiments show the benefits of our pre-trained model in reducing the dependency on large amount of task-specific labels.

We also point out several limitations of our method, such as no explicit IP node interaction modeling, and traffic feature discretization being a lossy process. Future works could address these limitations, and explore more varieties of networking tasks for fine-tuning, as well as building better benchmark data for developing foundation models in networks.

\minlan{If we have space, we can have a separate discussion section about these points... Maybe also argue more about the generality of your approach in incorporating other types of data than NetFlow data?}

\notes{
\vspace{0.1in}

Limitations:

- no nodes interactions during pre-training. can apply gnn objectives

- no micro-level inside packet learning. we are based on volumetric traffic (this is fine as we base on flow level)

- discretization loses some details in traffic volumes

}


\bibliographystyle{ACM-Reference-Format}
\bibliography{references}

\end{document}